\documentclass[11pt]{article}
\usepackage[utf8]{inputenc}
\usepackage[english]{babel}
\usepackage{amsmath, amssymb, amsthm}
\usepackage{graphicx}
\usepackage{url}
\usepackage{hyperref}
\usepackage{cite}
\usepackage{geometry}
\usepackage{booktabs}
\usepackage{array}
\usepackage{tabularx}
\usepackage{tikz}
\usepackage{xcolor}
\usepackage{algorithm}
\usepackage{algorithmicx}
\usepackage{algpseudocode} % или algorithmic
\usetikzlibrary{arrows.meta}

\title{A Triadic Suffix Tokenization Scheme for Numerical Reasoning}

\author{Olga Chetverina\thanks{Lecturer at MIPT, Moscow, Russia. Email: tst.chetverina@gmail.com. ORCID: \href{https://orcid.org}{0000-0003-4924-8130}. Previously archived at Zenodo: DOI 10.5281/zenodo.18999577. This research was conducted independently outside of any institutional assignments, and no institutional resources were used.}}

\begin{document}

\maketitle

\begin{abstract}
Standard subword tokenization methods fragment numbers inconsistently, causing large language models (LLMs) to lose positional and decimal structure - a primary driver of errors in arithmetic and scientific reasoning.

We introduce Triadic Suffix Tokenization (TST), a deterministic scheme that partitions digits into three-digit triads and annotates each triad with an explicit magnitude marker. Critically, the scheme defines a fixed, one-to-one mapping between suffixes and orders of magnitude for the integer part (thousands, millions, billions, etc.) and a parallel system of replicated markers for fractional depth (tenths, thousandths, millionths, etc.). Unlike approaches that rely on positional inference, this method provides a consistent gradient signal, which should ensure stable convergence.

Two implementation variants are proposed: (1) a vocabulary-based approach that adds at most 10,000 fixed tokens to an existing vocabulary, covering 33 orders of magnitude ($10^{-15}$ to $10^{18}$); and (2) a suffix-marker approach that uses a small set of special tokens to denote magnitude dynamically. Both variants preserve exact digits while making order-of-magnitude relationships transparent at the token level. While we focus on 3-digit groups (Triadic), the framework is inherently scalable to any group size for precise vocabulary optimization. Furthermore, it allows for linear vocabulary expansion to accommodate arbitrary precision and range. TST is architecture-agnostic and can be integrated as a drop-in preprocessing step. Experimental validation is deferred to future work.
\end{abstract}

\section{Introduction}

Large Language Models (LLMs) have achieved remarkable performance on complex reasoning tasks, yet they frequently stumble on basic numerical understanding. The famous "$9.11>9.9$" failure exemplifies how models misjudge decimal magnitudes \cite{yang2024number}. This weakness stems largely from tokenization: numbers are fragmented into arbitrary subword units, losing their positional and magnitude information \cite{singh2024tokenization, thawani2022estimating}. Standard tokenizers split “100400” into “100” and “400” without encoding that the former represents hundreds of thousands. As a result, models must learn magnitude relationships from scratch — a statistically inefficient task.

Recent work has explored various solutions: specialized number tokens \cite{loukas2025system}, continuous number encodings such as xVal \cite{xval2023}, right‑to‑left tokenization with comma separation \cite{zhou2024scaling}, and comprehensive benchmarks like NumericBench \cite{li2025exposing} and Number Cookbook \cite{yang2024number}. Studies comparing base‑10 (digit‑level) versus base‑1000 (triad‑level) tokenization show that digit‑level approaches are more data‑efficient, but struggle with magnitude comprehension \cite{zhou2024scaling, singh2024tokenization}. 

This paper introduces a simple yet effective tokenization scheme that combines triad grouping (base‑1000) with explicit magnitude annotations for both integer and fractional parts, providing a stronger inductive bias for numerical reasoning.

\section{Related Work}

Digit-level tokenization (base-10) treats each digit as an independent token. While data-efficient, it lacks explicit magnitude cues, forcing models to infer order from positional patterns alone \cite{zhou2024scaling}. Multi-digit tokenization, used by GPT-3.5 and GPT-4, reduces sequence length but introduces arbitrary boundaries that can fragment numbers inconsistently \cite{singh2024tokenization}. 

xVal encodes numbers as continuous embeddings using a single token per number, demonstrating strong interpolation performance \cite{xval2023}. However, it discards exact digits, trading precision for smoothness—a trade-off unsuitable for tasks requiring exact arithmetic.

Right-to-left tokenization, which adds commas every three digits from the right, has been shown to improve integer arithmetic by up to 20\% compared to unformatted numbers \cite{singh2024tokenization, zhou2024scaling}. This demonstrates that even simple formatting shifts help models reason about numbers. However, commas only group digits; they do not indicate the magnitude of each group. A model still must infer whether “123” stands for 123, 123,000, or 123,000,000 from its position in the sequence.

Recent work has also explored modifying the loss function rather than tokenization. Number Token Loss (NTL) \cite{icml2025ntl} replaces the standard cross-entropy loss with a regression-like objective for numerical tokens, treating numbers as continuous values during training. This approach preserves exact digits while providing a smooth gradient signal. Importantly, NTL operates at the training level and is orthogonal to tokenization schemes—it can be combined with any input representation.

Several established benchmarks can assess numerical reasoning in future studies: NumericBench evaluates six fundamental numerical capabilities \cite{li2025exposing}, Number Cookbook covers 17 distinct numerical tasks \cite{yang2024number}, and studies probing enumeration skills show that even state-of-the-art models lack systematic enumeration \cite{daibasoglu2025probing}.

\section{Proposed Method: Triadic Suffix Tokenization (TST)}

\subsection{Core Principles}
\begin{enumerate}
    \item Group digits into triads (thousands, millions, etc.).
    \item Annotate each triad with explicit magnitude markers.
    \item Preserve exact digits.
\end{enumerate}

\subsection{Integer Part}
Digits are grouped from right to left:
\begin{align*}
100400 &\rightarrow 100\text{k}\,400 \\
1234567 &\rightarrow 1\text{m}\,234\text{k}\,567 \\
100200400 &\rightarrow 100\text{m}\,200\text{k}\,400 \\
123456789012345 &\rightarrow 123\text{t}\,456\text{b}\,789\text{m}\,012\text{k}\,345
\end{align*}

\begin{table}[h]
\centering
\begin{tabular}{|c|c|c|}
\hline
\textbf{Suffix} & \textbf{Magnitude} & \textbf{Power} \\
\hline
k & thousand & \(10^3\) \\
m & million & \(10^6\) \\
b & billion & \(10^9\) \\
t & trillion & \(10^{12}\) \\
q & quadrillion & \(10^{15}\) \\
\hline
\end{tabular}
\caption{Integer suffixes}
\end{table}

\subsubsection{Why Suffixes, Not Just Commas}
Unlike right-to-left commas, which only group digits, the suffixes k, m, b, t, q explicitly indicate the order of magnitude. This gives the model direct access to the scale of each digit group, rather than forcing it to infer magnitude from position.

\subsection{Fractional Part}
Fractional digits are grouped left to right with replicated 'p' markers. To ensure a canonical representation and maintain a 1:1 mapping between tokens and numerical values, all fractional triads are \textbf{right-padded with trailing zeros} to a fixed length of three digits. This normalization ensures that numerically equivalent representations (e.g., $0.1$ and $0.100$) result in the same token sequence.

\begin{align*}
1.12345678 &\rightarrow 1.\ 123p\ 456pp\ 780ppp \\
0.0045 &\rightarrow 0.\ 004p\ 500pp \\
\pi &\rightarrow 3.\ 141p\ 592pp\ 653ppp\ 589pppp\ 793ppppp
\end{align*}

Maximum practical depth is typically 5 'p's (15 decimal places). Without padding, the vocabulary would unnecessarily expand to include sub-triadic variations, and the model would be forced to learn that $0.1, 0.10,$ and $0.100$ represent the same value independently. Padding ensures that each token consistently represents a value in the form $n \times 10^{-3k}$ where $n \in [0, 999]$, requiring only 1,000 tokens for each triad depth.

This normalization prioritizes numerical consistency and computational precision over the preservation of the original surface form. While mapping numerically equivalent values to a single sequence improves convergence, some tasks require trailing zeros to carry semantic meaning. A detailed discussion of this trade-off and a proposed minor extension for precision-preserving applications are provided in Section~\ref{sec:limitations}.

\subsection{Vocabulary Considerations}
Two options are possible:

\begin{itemize}
    \item \textbf{Option A (Separate tokens):} Keep digit groups and suffixes as separate tokens. Vocabulary adds only ~10 new tokens (k, m, b, t, q, p, pp, ppp, pppp, ppppp). This minimally expands the vocabulary; models learn to combine digits with suffixes through attention.
    \item \textbf{Option B (Compound tokens):} Create combined tokens like “100k”, “234m”. With triads from 000--999 and 5 integer suffixes: \(1000 \times 5 = 5000\) tokens. For fractions with up to 5 'p' depths and triads 000--999: another 5000 tokens. Total ~10,000 new tokens — manageable for modern vocabularies.
\end{itemize}

Option B produces shorter input sequences and presents the model with ready-made magnitude-digit units, eliminating ambiguity about which suffix belongs to which digit group. While the smaller vocabulary of Option A is appealing, the trade-off remains an empirical question.

While we focus on $N=3$ (Triadic) for its alignment with human readability and its effective balance between vocabulary size and sequence length, the framework is fundamentally designed for any group size $N$, allowing for precise optimization of the model's vocabulary footprint.  Generally, Option A is more practical for larger group sizes ($N \geq 3$) where the relative token overhead is negligible.

As for Option B, even in the case of $N=1$, the input context length remains no larger than that of standard digit-by-digit representations. We hypothesize that the \textbf{$N=1$ configuration} is optimal for \textbf{Small Language Models (SLMs)}. In this setup, the representation is extremely compact (e.g., only 330 additional tokens to cover a range up to $10^{33}$), and the bijective nature of the tokens eliminates ``semantic blur". By providing an explicit one-to-one mapping between each digit and its magnitude, $N=1$ reduces the representational burden on constrained weight matrices, potentially leading to more robust arithmetic grounding in sub-billion parameter architectures.

The complete deterministic procedure for the TST scheme, illustrating the generation of magnitude-aware compound tokens as described in Option B, is formalized in Algorithm~\ref{alg:tst}. For Option A, the algorithm remains identical except in Step 4, where the digit group and the suffix are appended as two separate tokens instead of a single compound unit.

\begin{algorithm}[H]
\caption{Generalized Suffix Tokenization (GST) Mapping}
\label{alg:tst}
\begin{algorithmic}
\Require{A raw numerical string $S$, and internal group size $N$ optimized for the model's vocabulary.}
\Ensure{A sequence of magnitude-aware tokens $T$}

\Statex \textbf{// Step 1: Normalization and Cleansing}
\State $\text{Prefix} \gets \text{ExtractPrefix}(S)$ 
\State $S' \gets \text{NormalizeToStandardDigits}(S)$ 
\State $I, F \gets \text{SplitIntoIntegerAndFractionalParts}(S')$

\Statex \textbf{// Step 2: Integer Processing}
\State $\mathcal{I} \gets \text{DivideIntoGroupsOfSizeFromRightToLeft}(I, N)$
\State $I' \gets \text{PadWithZerosFromLeftToMultipleOf}(\mathcal{I}, N)$

\Statex \textbf{// Step 3: Canonical Fraction Processing}
\State $\mathcal{F} \gets \text{DivideIntoGroupsOfSizeFromLeftToRight}(F, N)$
\State $F' \gets \text{PadWithZerosFromRightToMultipleOf}(\mathcal{F}, N)$

\Statex \textbf{// Step 4: Tokens Generation with Suffixes}
\State $T \gets [\ ]$

\If{$\text{Prefix is not empty}$}
    \State $T.\text{append}(\text{Prefix})$
\EndIf

\Statex \textbf{// Process Integer Parts (Option B depicted)}
\For{\textbf{each} group $G_k$ \textbf{in} $\mathcal{I}$ at magnitude order $k$}
    \State $\text{suffix} \gets \text{GetMagnitudeSuffix}(k)$ 
    \State $T.\text{append}(G_k + \text{suffix})$ 
\EndFor

\State $T.\text{append}(\text{"."})$

\Statex \textbf{// Process Fractional Parts}
\For{\textbf{each} group $G_p$ \textbf{in} $\mathcal{F}$ at decimal depth $p$}
    \State $\text{suffix} \gets \text{GetPrecisionSuffix}(p)$
    \State $T.\text{append}(G_p + \text{suffix})$
\EndFor

\State \Return $T$
\end{algorithmic}
\end{algorithm}

\section{Annotation Placement: Prefix vs. Suffix}

A related line of work, NumeroLogic \cite{numerologic2024}, adds a prefix indicating the total number of digits in a number (e.g., \texttt{<sn>2<mn>42<en>} for 42). On short-number arithmetic (addition of numbers up to three digits), this approach improves accuracy, demonstrating that explicit structural annotations help. However, NumeroLogic provides a single prefix per number and does not annotate individual triads. Its effectiveness on longer numbers or fractional precision tasks remains an open question.

TST, in contrast, is designed to scale across 33 orders of magnitude (\(10^{-15}\) to \(10^{18}\)) by annotating each triad with its magnitude. Given the success of prefix-based length annotation on short numbers, one might expect similar or stronger benefits from TST's more detailed per-triad annotation across the full numeric range.

Beyond the choice between a single length prefix and per-triad suffixes, there is also the question of where to place the marker within each triad: before the digits (prefix) or after (suffix). For example, the number 123,456 could be represented as:

\begin{itemize}
    \item \textbf{Suffix format}: \texttt{123k 456}
    \item \textbf{Prefix format}: \texttt{k123 456}
\end{itemize}

When using compound tokens (Option B: \texttt{123k} or \texttt{k123} as a single token), the distinction between prefix and suffix disappears at the token level—both become atomic vocabulary units, and the embedding layer treats them identically regardless of internal order.

However, when numbers are tokenized digit by digit, suffix placement offers several advantages:

\begin{itemize}
    \item \textbf{Deterministic boundaries}: The suffix marks the end of a triad. For \texttt{1 2 3k 4}, the model knows that “1 2 3” form a complete thousand group before seeing further digits. With a prefix (\texttt{k 1 2 3 4}), the model cannot know where the group ends.
    \item \textbf{Alignment with human reading}: Suffix notation (e.g., “123k”) matches the natural way numbers are written with separators (123,000) and spoken (“one hundred twenty-three thousand”).
    \item \textbf{Implicit length information}: A suffix like \texttt{q} (quadrillion) tells the model that exactly 15 more digits (five triads) follow, providing a strong prior for the total length of the integer part.
\end{itemize}

Prefix placement could potentially provide the magnitude information slightly earlier, but it does so at the cost of slight boundary ambiguity. For these reasons, we adopt the \textbf{suffix} placement for TST when digit-level tokenization is used. In the compound-token variant (Option B), the choice is immaterial. The suffix versus prefix question remains an empirical one for some cases, but the deterministic boundary property makes suffix a natural and robust default.

\section{Statistical Analysis}

\subsection{Comparison Framework}

The comparison of different approaches is provided in the Table~\ref{tab:comparison}.
\begin{table}[htbp]
\centering
\caption{Comparison of tokenization schemes: inductive biases.}
\label{tab:comparison}
\begin{tabular}{l	cp{2.5cm}p{2cm}>{\raggedright\arraybackslash}p{3.8cm}}
\toprule
\textbf{Scheme} & \textbf{Exact digits} & \textbf{Magnitude info} & \textbf{Sequence length} & \textbf{Inductive bias} \\
\midrule
Digit-level (base-10)      & $\checkmark$ preserved & Implicit (must learn) & Long        & Digits independent \\[20pt] xVal~\cite{xval2023}       & $\times$ lost          & Continuous encoding   & 1 token \newline
  per number & Smoothness \\[20pt] 
Right-to-left~\cite{singh2024tokenization} & $\checkmark$ preserved & Positional only       & Medium      & Place-value alignment \\[12pt] 
\textbf{TST (proposed)}    & $\checkmark$ preserved & $\checkmark$ Explicit  & Medium      & Explicit hierarchy \\[12pt] \bottomrule
\end{tabular}
\end{table}

\subsection{Flexibility: Suffixes with Digit-Level Tokenization}

Even if one chooses to tokenize numbers as individual digits (base-10), suffixes can still provide magnitude cues without sacrificing digit-level precision. Consider the number 1,234,567:

\begin{itemize}
    \item \textbf{Pure digit-level}: \texttt{1 2 3 4 5 6 7} — no magnitude information.
    \item \textbf{Digit-level with suffixes}: \texttt{1m 2 3 4k 5 6 7} — explicit magnitude hints for each triad, preserving exact digits while adding structural cues.
\end{itemize}

This hybrid approach increases sequence length compared to triad-level grouping but offers a flexible trade-off: the model receives exact digits (preserving precision) alongside explicit magnitude annotations (improving reasoning). Suffixes thus enable designers to choose between compact sequences (triad-level) or maximal digit-level precision with structural hints—a flexibility not offered by pure digit-level or pure group-level schemes.

\subsection{Hypothesis}
Explicit magnitude annotation provides a stronger inductive bias than implicit or continuous approaches. TST preserves exact digits while making order explicit, potentially offering the best of both worlds. The suffix-based design also enables graceful trade-offs between sequence length and precision, depending on downstream requirements.

\section{Discussion}

\subsection{Advantages}

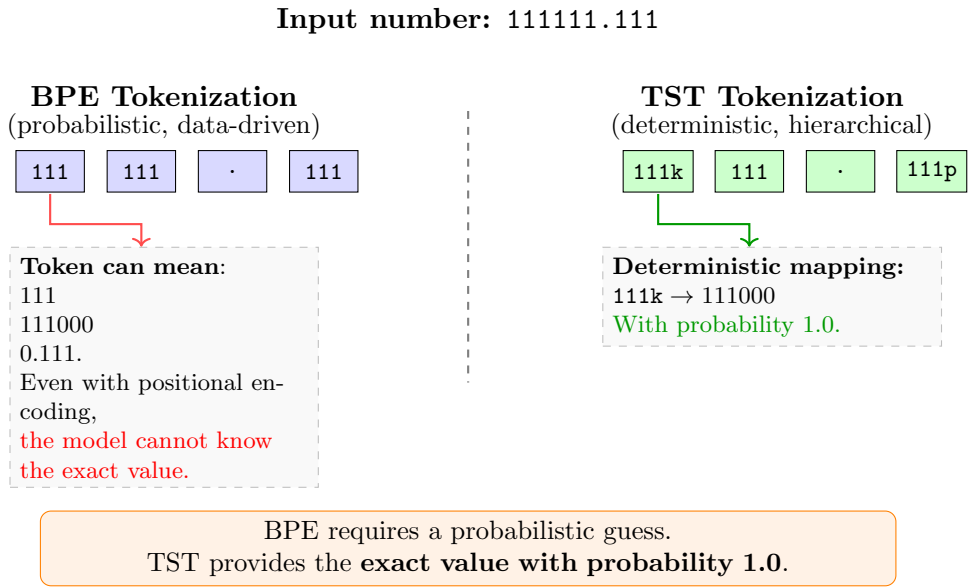
\begin{figure}[htbp]
\centering
\begin{tikzpicture}[
    token/.style={rectangle, draw, minimum width=0.9cm, minimum height=0.55cm, align=center, font=\ttfamily\footnotesize},
    note/.style={rectangle, draw=gray!50, dashed, fill=gray!5, align=left, font=\footnotesize},
    prob/.style={font=\tiny\itshape, text=red!70!black}
]

% ========== ВЕРХНИЙ БЛОК ==========
\node[draw=none, font=\bfseries] at (0,0) {Input number: \texttt{111111.111}};

% ========== BPE (ЛЕВАЯ КОЛОНКА) ==========
\node[font=\bfseries, align=center] at (-4,-1) {BPE Tokenization};
\node[font=\small, align=center] at (-4,-1.4) {(probabilistic, data-driven)};

% Токены BPE
\node[token, fill=blue!15] at (-5.5,-2) {\texttt{111}};
\node[token, fill=blue!15] at (-4.3,-2) {\texttt{111}};
\node[token, fill=blue!15] at (-3.1,-2) {\texttt{.}};
\node[token, fill=blue!15] at (-1.9,-2) {\texttt{111}};

% Стрелка
\draw[->, thick, red!70] (-5.5,-2.3) -- (-5.5,-2.7) -| (-4.3,-3);

% Вопрос
\node[prob] at (-5.5,-3.2) {?};

% Пояснение BPE
\node[note, text width=3.8cm, anchor=north] at (-4,-3) {
    \textbf{Token can mean}: \\
    $111$ \\ $111000$ \\ $0.111$. \\
    Even with positional encoding, \\
    \textcolor{red}{the model cannot know the exact value.}
};

% ========== РАЗДЕЛИТЕЛЬ ==========
\draw[gray, dashed, thick] (0,-1.2) -- (0,-4.8);

% ========== TST (ПРАВАЯ КОЛОНКА) ==========
\node[font=\bfseries, align=center] at (4,-1) {TST Tokenization};
\node[font=\small, align=center] at (4,-1.4) {(deterministic, hierarchical)};

% Токены TST
\node[token, fill=green!20] at (2.5,-2) {\texttt{111k}};
\node[token, fill=green!20] at (3.7,-2) {\texttt{111}};
\node[token, fill=green!20] at (4.9,-2) {\texttt{.}};
\node[token, fill=green!20] at (6.1,-2) {\texttt{111p}};

% Стрелка
\draw[->, thick, green!60!black] (2.5,-2.3) -- (2.5,-2.7) -| (3.7,-3);

% Пояснение TST

\node[note, text width=4.2cm, anchor=north] at (4,-3) {
    \textbf{Deterministic mapping:} \\
    $\texttt{111k} \rightarrow 111000$ \\

    \textcolor{green!60!black}{With probability 1.0.}

};

% ========== ВЫВОД ==========
\node[draw=orange, fill=orange!10, rounded corners, text width=11cm, align=center, font=\small] at (0,-7) {
    BPE requires a probabilistic guess. \\
    TST provides the \textbf{exact value with probability 1.0}.
};

\end{tikzpicture}
\caption{Comparison: BPE (left) vs TST  Option B  (right) for tokenizing 111111.111.}
\label{fig:bpe_tst_comparison}
\end{figure}

The following advantages primarily reflect Option B representation, though many also apply  to Option A features.

\begin{itemize}
    \item \textbf{Explicit real token value}: As shown in Figure~\ref{fig:bpe_tst_comparison} unlike other approaches, TST in option B establishes a bijective mapping between the token and its real numerical value, ensuring no ambiguity in decoding.
    
    \item \textbf{Fractional precision}: It guarantees that the model understands the decimal part since replicated 'p' markers create an ordinal scale of decimal depth. 
    
    \item \textbf{Compatibility}: TST requires only a different tokenizer; no modifications to the model architecture are needed. While new token embeddings must be learned, the core model remains unchanged. Moreover, for Option B (compound tokens), the choice of suffix symbol is immaterial — each combined token receives its own embedding, eliminating any risk of linguistic conflict (e.g., between suffix ``k'' and the letter ``k'' in text). Option A may require reserved tokens as $[MAG_k], [MAG_p]$  etc. to avoid such conflicts.
    
    \item \textbf{Complementarity with NTL}: TST operates at the input level (tokenization), while NTL \cite{icml2025ntl} modifies the loss function to incorporate numerical closeness. The two are orthogonal and can be combined: TST provides explicit structural information at the input, and NTL adds a numerical inductive bias during training. Their synergy may yield stronger improvements than either approach alone.
 
    \item \textbf{Stable convergence}: We hypothesize that explicit real token values provide a consistent gradient signal, leading to faster and more stable convergence. By removing the need for the model to infer numerical magnitude from context, TST would likely reduce both inference errors and the computational cost of training.
    
    \item \textbf{Scalability to arbitrary precision}: The TST method's current 33-order magnitude range is fully scalable, allowing for extension to arbitrary ranges by defining additional suffix tokens without changing the core triadic logic. This flexible token budget allows the framework to adapt to specific domain requirements, spanning from quantum physics to astronomy, simply by adjusting the vocabulary size. Extending TST to cover an additional three orders of magnitude requires exactly 1000 new tokens — all triads 000--999 with the new suffix.     
    
    \item \textbf{Canonical fractions}: Zero padding ensures that $0.1$, $0.10$, and $0.100$ all map to the same token sequence $0.\ 100p$, unlike standard tokenization where they differ. This eliminates surface-form ambiguity for numerically equivalent numbers. For example, even for Option A, comparison of $0.19 \rightarrow 190 p$ and $0.9 \rightarrow 900 p$ becomes obvious.    
\end{itemize}

\subsection{Limitations and Future Work}
\label{sec:limitations}

\begin{itemize}
    \item \textbf{Experimental validation}: This paper establishes the theoretical framework and algorithmic foundation for TST.  Although in a production environment (Option B) numbers are mapped directly to magnitude-aware tokens, the suffix-based notation is effective during the experimental stage. It allows for a modular testing workflow: a researcher can implement the parser separately and simply expand the model's vocabulary, evaluating the scheme’s impact without deep modifications to the underlying tokenization engine. A reference implementation of the TST scheme  for both Option A and Option B and the resulting vocabulary for Option B are provided for reproducibility at \cite{tst_github}. It is intended to demonstrate the triadic suffix mapping and is not yet an exhaustive library for all numerical notation. Future work should evaluate on numerical reasoning benchmarks like NumericBench \cite{li2025exposing} and arithmetic tasks, comparing against digit-level, xVal, right-to-left \cite{singh2024tokenization, zhou2024scaling}, and NumeroLogic \cite{numerologic2024}. 
    
    \item \textbf{Tokenization determinism}: Numbers written without separators are processed unambiguously: integer parts are grouped from the right, fractional parts from the left. Numbers with fewer than four digits receive no suffix. This deterministic procedure works for any numeric string and requires no complex heuristics.

\item \textbf{Padding fractional parts:} 
Normalizing fractional parts to three digits discards trailing zeros, mapping numerically equivalent values (e.g., $0.1, 0.10, 0.100$) to a single token sequence (\texttt{100p}). For tasks where the original surface form carries semantic meaning --- such as fixed-point financial data or significant figures in scientific records --- this loss of representational information may be undesirable. 

To address this, the canonical 1,000-token scheme can be augmented with length-specific terminator tokens: $0.1 \rightarrow \texttt{0. 100p [T1]}$, and $0.10 \rightarrow \texttt{0. 100p [T2]}$. In this case, the numerical part remains identical, eliminating the need to expand the vocabulary with additional sub-triadic tokens (\texttt{0p}--\texttt{99p}). The meta-token explicitly indicates the original format. Since this approach slightly increases the sequence length, it is best reserved as an optional mode for domain-specific applications where original formatting is critical.

    \item \textbf{Alternative grouping sizes}: While the 3-digit grouping (base-1000) is natural for human readability, the framework allows for any $N$. For instance, Option B with $N=1$ might be more effective for SLMs, but other sizes could also be considered. Language-specific grouping (e.g., 3-2-2 in India, 4-4 in China) can be handled by the parser without changing the core vocabulary \cite{tst_github}. The final decision should be guided by which representation is most efficient for the specific model architecture.    

    \item \textbf{Tokenisation variants}: Option B (compound tokens) produces shorter input sequences (lower \emph{token count}) and gives the model ready‑made magnitude‑digit units, eliminating any ambiguity about which suffix belongs to which digit group. Option A (separate suffixes) adds only ~10 new tokens. The trade-off between sequence length and vocabulary size remains an empirical question.

    \item \textbf{Comparison with xVal and NTL}: xVal demonstrates strong interpolation performance \cite{xval2023}; NTL offers a complementary training-level approach \cite{icml2025ntl}. Their combination with TST may yield synergistic improvements.
\end{itemize}

\section{Conclusion}
We present Triadic Suffix Tokenization (TST), a scheme that transforms numerical tokenization from a source of errors into a structured representation. Unlike right-to-left commas, which only group digits, TST gives the model explicit knowledge of magnitude for integers and the same structural clarity for fractions. Our analysis compared TST with existing approaches, including digit-level tokenization, xVal, right-to-left commas, NumeroLogic, and NTL. While commas provide grouping without magnitude, and digit count prefixes improve accuracy on short numbers, none offer explicit per-triad magnitude annotation or deterministic boundary resolution. TST addresses these limitations through a fixed bijective mapping between suffixes and orders of magnitude, replicated markers for fractional depth, and implicit length information for integer parts.

The theoretical advantages of TST — explicit hierarchy, deterministic boundaries, and scalability across 33 orders of magnitude (\(10^{-15}\) to \(10^{18}\)) — suggest that it may outperform existing methods across the full numeric range. While we focus on $N=3$ for human readability, the scheme is fundamentally extensible to any group size $N$ and can be expanded indefinitely by introducing additional suffix markers. By removing tokenization ambiguity and ensuring canonical representation through zero padding, the scheme ensures that the model receives an unambiguous gradient signal for numerical values, which may reduce both inference errors and training cost. The scheme requires no architectural modifications, only tokenization preprocessing, making it a practical drop-in enhancement. Moreover, TST is orthogonal to training-level improvements such as NTL; their combination may yield synergistic benefits.

Empirical validation on numerical reasoning benchmarks such as NumericBench~\cite{li2025exposing} (comparing TST against digit-level tokenization, xVal~\cite{xval2023}, right-to-left commas~\cite{singh2024tokenization, zhou2024scaling}, and NumeroLogic~\cite{numerologic2024}) remains for future work. If confirmed, TST could offer a simple yet powerful enhancement to any language model that must reason about numbers.

\end{document}